\documentclass[sigconf,natbib=true]{acmart}
\usepackage{kotex}

\usepackage{multirow, multicol, caption}
\usepackage{array}
\newcolumntype{L}[1]{>{\raggedright\arraybackslash}p{#1}}
\usepackage{adjustbox}
\usepackage[ruled,noend]{algorithm2e}
\DontPrintSemicolon
\usepackage{algorithmic}
\usepackage{bm}
\usepackage[utf8]{inputenc}
\usepackage{subcaption}
\usepackage{booktabs}
\usepackage{siunitx}
\usepackage{rotating}
\usepackage{colortbl}
\usepackage{xcolor}
\usepackage{enumitem}
\setlist[itemize]{leftmargin=*, topsep=2pt, itemsep=1pt, parsep=0pt, partopsep=0pt}
\usepackage{makecell}

\usepackage{amsmath}

\usepackage{amssymb}
\usepackage{amsthm}

\theoremstyle{definition}

\usepackage{tablefootnote}
\usepackage{graphicx}
\usepackage{placeins}
\usepackage{xurl}
\usepackage{xspace}
\usepackage{tikz}
\usetikzlibrary{arrows.meta, positioning, fit, backgrounds, shapes.geometric, decorations.pathreplacing}
\definecolor{vdark}{HTML}{3B528B}
\definecolor{vteal}{HTML}{21918C}
\definecolor{vgreen}{HTML}{5EC962}
\definecolor{vlime}{HTML}{BDDF26}
\newcommand{\model}[1]{\texttt{\small\nolinkurl{#1}}}
\newcommand{\method}{\texttt{TRACK}\xspace}
\newcommand{\hofinet}{HOFINET\xspace}
\newcommand{\score}[2]{\shortstack{#1\\[-2pt]{\tiny $\pm\,#2$}}}
\usepackage{pifont}
\newcommand{\cmark}{\ding{51}}
\newcommand{\xmark}{\ding{55}}

\usepackage[disable,textsize=tiny]{todonotes}
\usepackage{marginnote}
\let\marginpar\marginnote
\usepackage[T1]{fontenc}

\AtBeginDocument{}

\setcopyright{none}

\acmConference[WSDM '27]{The 20th ACM International Conference on Web Search and Data Mining}{February 15--19, 2027}{Hong Kong SAR, China}
\acmBooktitle{Proceedings of the Twentieth ACM International Conference on Web Search and Data Mining (WSDM '27), February 15--19, 2027, Hong Kong SAR, China}
\acmYear{2027}
\copyrightyear{2027}
\acmDOI{}
\acmISBN{}

\begin{document}

\title[TRACK: Multiple Instance Learning for AML]{TRACK: Multiple Instance Learning for Anti-Money Laundering in Transaction Networks}

\author{Seonkyu Lim}
\orcid{0000-0003-4904-7144}
\authornote{These authors contributed equally to this work.}
\affiliation{
  \institution{KFTC}
  \city{Seoul}
  \country{South Korea}
}
\email{sklim@kftc.or.kr}

\author{Inwoo Tae}
\orcid{0000-0001-8087-8240}
\authornotemark[1]
\affiliation{
  \institution{UNIST}
  \city{Ulsan}
  \country{South Korea}
}
\email{vitainu0104@unist.ac.kr}

\author{Gwangui Hong}
\orcid{0009-0003-9160-6699}
\authornotemark[1]
\affiliation{\institution{KAIST}\city{Daejeon}\country{South Korea}}
\affiliation{\institution{KFTC}\city{Seoul}\country{South Korea}}
\email{recordame@kftc.or.kr}

\author{Jaehoon Lee}
\orcid{0009-0002-0744-6593}
\affiliation{
  \institution{LG AI Research}
  \city{Seoul}
  \country{South Korea}
}
\email{jaehoon.lee@lgresearch.ai}

\author{Jeongwhan Choi}
\orcid{0000-0002-6530-2662}
\affiliation{
  \institution{KAIST}
  \city{Daejeon}
  \country{South Korea}
}
\email{jeongwhan.choi@kaist.ac.kr}

\author{Taehoon Kim}
\orcid{0000-0003-4788-4651}
\authornote{Corresponding authors}
\affiliation{
  \institution{Sogang University}
  \city{Seoul}
  \country{South Korea}
}
\email{taehoonkim@sogang.ac.kr}

\author{Min-Soo Kim}
\orcid{0000-0002-5065-0226}
\authornotemark[2]
\affiliation{
  \institution{KAIST}
  \city{Daejeon}
  \country{South Korea}
}
\email{minsoo.k@kaist.ac.kr}

\author{Yongjae Lee}
\orcid{0009-0006-9807-0198}
\authornotemark[2]
\affiliation{
  \institution{UNIST}
  \city{Ulsan}
  \country{South Korea}
}
\affiliation{
  \institution{LinqAlpha}
  \city{New York}
  \country{USA}
}
\email{yongjaelee@unist.ac.kr}

\renewcommand{\shortauthors}{Lim et al.}

\begin{abstract}
Anti-money laundering (AML) investigations are organized around accounts, while the suspicious evidence for an account may reside in only a few of its transactions. Yet many account-level detectors summarize transactions before learning, and message-passing models may aggregate once more over neighboring accounts. We name this discrepancy the aggregation mismatch and study it under an account-window benchmark label defined by the presence of at least one flagged transaction. We recast account-level AML detection as multiple instance learning, in which an account-window is a bag and its transactions are instances. We propose \method, which encodes every transaction with a shared encoder and selectively pools the encodings into an account decision. To emulate coarse supervision, we withhold the transaction labels of five
AML datasets from \method throughout training. Under a time-ordered protocol, \method achieves the highest average precision among the tabular and graph baselines trained from account labels on every dataset. Despite using only account-window supervision, it remains competitive with edge classifiers trained directly from transaction labels. Its instance scores also rank flagged transactions well above chance and approach or exceed directly supervised ranking performance across datasets.
\end{abstract}



\maketitle

\section{Introduction}\label{sec:intro}

\begin{figure*}[t]
\centering
\includegraphics[width=\textwidth]{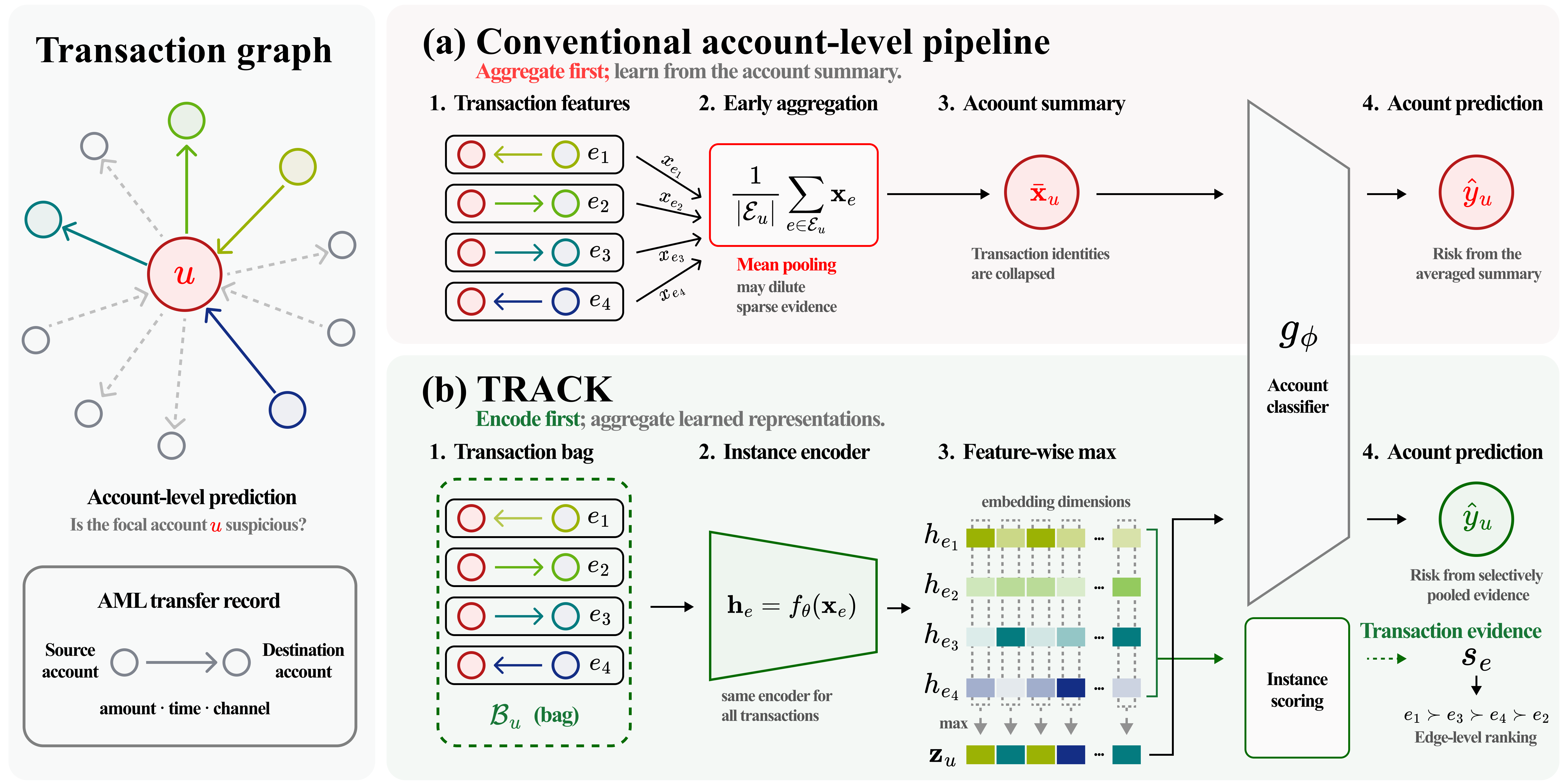}
\caption{Overview of the aggregation mismatch and \method. (a) Aggregate-first modeling averages transaction features before learning. (b) \method encodes transactions as instances and applies feature-wise max pooling, yielding account risk and transaction ranking from account labels alone.}
\label{fig:overview}
\end{figure*}

Money laundering is estimated to account for 2--5\% of global GDP, posing a substantial threat to the integrity of financial systems~\cite{unodc_ml}. To combat this threat, financial institutions operate anti-money laundering (AML) systems that monitor massive volumes of transactions for suspicious activity. Laundering activity may involve patterns of transfers across multiple accounts and over time~\cite{fatf2018pml,li2020flowscope}. This relational and temporal structure can be naturally represented as a transaction graph, where accounts are nodes and transactions are directed, attributed edges.

This graph representation provides financial institutions with a natural basis for AML monitoring and investigation. However, financial institutions process vast numbers of transactions every day, making it impractical for investigators to examine each transaction individually. AML teams therefore prioritize suspicious customers or accounts and then review their transactions and connected accounts~\cite{fatf2013guidance}. In this process, accounts are the primary targets of investigation, while transactions provide the supporting evidence. Accordingly, we formulate AML detection as account-level node classification rather than transaction-level edge classification.

However, many public and synthetic AML benchmarks provide transaction-level labels~\cite{amlpatterns2024realistic,huda2025amlnet,suzumura2021amlsim, elmougy2023demystifying}. As a result, a line of work formulates AML detection as transaction-level edge classification and evaluates models directly on labelled transfers. This formulation is appropriate when the goal is to screen or block an individual transaction. Transaction-level performance, however, does not by itself measure whether a model prioritizes the customer or account cases that analysts investigate. Moreover, supervised edge classification assumes that fine-grained transaction labels are available for training. It therefore answers a different operational question from the account-centered setting we target.

Account- and node-level methods use the appropriate decision unit, but a common pipeline compresses each account's transactions into a fixed-length summary before learning. Tabular models classify aggregated statistics of transaction activity, while graph models may apply message passing on top of similar node representations~\cite{dou2020enhancing,liu2021pick,tang2022rethinking,wang2023label}. However, the evidence may be concentrated in only a small subset of an account's transactions. Aggregating many transactions can dilute that signal and obscure which transactions support the account-level prediction. We call this discrepancy the \emph{aggregation mismatch}: the decision is made at the node level using evidence from transaction edges, but those edges are collapsed before the supervised model can learn from them.

We propose \textbf{T}rans\-ac\-tion \textbf{R}ep\-re\-sen\-ta\-tion for \textbf{A}c\-count \textbf{C}las\-si\-fi\-ca\-tion and evidence ran\textbf{K}ing (\method), a transaction-preserving node classifier joining node decisions to transaction evidence. \method separates the unit of decision from the unit of evidence: it predicts whether an account is suspicious while learning from its individual transactions. To achieve this, \method represents each account-window as a bag of transactions under the multiple-instance learning framework~\cite{dietterich1997solving,ilse2018attention}. Unlike aggregate-first account pipelines, \method keeps transactions as instances until their representations are learned, then combines them. The learned embeddings support a complete candidate ranking, while the max-pooled dimensions provide a separate exact attribution of the account logit.

We evaluate this ac\-count-su\-per\-vised setting by coarsening the transaction labels in AML benchmarks into ac\-count-win\-dow labels and with\-hold\-ing the original flags throughout training. The hidden labels are used only for trans\-ac\-tion-rank\-ing eval\-u\-a\-tion, controlled diagnostics, and transaction-supervised baselines. This controlled construction makes the MIL relation exact; it does not model institutional cases whose labels arise only from higher-order patterns, external information, reporting delay, or investigator judgment. Under a time-ordered protocol, we compare \method with tabular, graph, and transaction-supervised alternatives and isolate the effect of aggregation location and operator. Our contributions are:

\begin{itemize}[leftmargin=*, topsep=2pt, itemsep=1pt, parsep=0pt, partopsep=0pt]
\item We identify the aggregation mismatch in account-level AML: aggregate-first node classifiers hold the right prediction target but collapse transaction identity and attribute co-occurrence.
\item We recast account-level detection as multiple instance learning, in which an
account-window is a bag and its transactions are instances, aligning the aggregation
the model performs with the rule that assigns the label.
\item We propose \method, a transaction-preserving node classifier that learns from account labels, pools transaction representations selectively, and provides candidate ranking and forward-path attribution without graph message passing.
\item We construct a controlled account-supervision benchmark by coarsening transaction labels and withholding them from \method training, then assess account detection, within-account transaction ranking, and aggregation behavior against tabular, graph, and transaction-supervised alternatives.
\end{itemize}

\section{Related Work}\label{sec:related_work}

\subsection{Account-Level Detection}
Graph-based account-level detection has largely developed around the context surrounding a target node. Earlier studies focused on making message propagation robust to camouflage, label imbalance, unreliable neighbors, and the low homophily of fraudulent networks~\cite{dou2020enhancing,liu2021pick,peng2021reinforced,liu2020alleviating, zhang2021fraudre,tang2022rethinking,wang2023label,chen2024consisgad}. AML research has used transaction-graph structure to learn representations and find suspicious flow patterns under limited supervision~\cite{cardoso2022laundrograph,lu2024graph,li2020flowscope}. Recent methods enlarge or refine the graph context used for node decisions. GCD-GNN estimates class-level prototypes and uses global confidence to control how information from different types of neighbors is aggregated~\cite{liu2025gcd}. GPA instead encodes variable-length paths, enabling a node classifier to capture associations beyond its immediate neighborhood~\cite{qiu2025gpa}. In contrast, \method learns account-level decisions directly from a set of transaction instances under account-level supervision.

\begin{table*}[t]
\centering
\caption{Transaction-instance features used by \method. Each instance contains six transaction attributes and $22$ history features for the focal account, with the latter repeated for the counterparty (up to $50$ in total). Missing channel or clock fields reduce it.}
\label{tab:featurelist}
\small
\setlength{\tabcolsep}{6pt}
\begin{tabular}{@{}L{0.15\textwidth}cL{0.77\textwidth}@{}}
\toprule
Group & $n$ & Feature definitions \\
\midrule
Transfer & 6 & Log amount; direction; cyclic hour encoding; two channel indicators \\
Amount & 4 & Amount z-score and percentile rank; share of cumulative volume; frequency of similar prior amounts \\
Rhythm & 5 & Time since the previous transfer and its z-score; first-transaction indicator; prior transfer count; account tenure \\
Counterparty & 3 & New-counterparty indicator; prior transaction count and elapsed time for the account pair \\
Balance & 2 & Normalized outgoing--incoming differences in transaction count and amount \\
Pass-through & 2 & Time since the most recent incoming transfer; current-to-incoming amount ratio \\
Windows & 6 & Log transaction count and new-counterparty rate over three trailing horizons ($1/6$, $0.4$, and $0.8$ of the observed span) \\
\bottomrule
\end{tabular}
\end{table*}

\subsection{Transaction-Level Detection}
Recent graph-based AML studies, particularly those evaluated on transaction-level benchmarks, have predominantly formulated detection as edge classification, treating each transaction as the prediction target~\cite{egressy2024directed,lin2024fraudgt}. Across this line of work, methods have sought to capture the directionality and multiplicity of transactions~\cite{egressy2024directed,bilgi2024mega}, transaction attributes~\cite{lin2024fraudgt}, temporal dependencies~\cite{gounoue2025temptrag}, and higher-order transaction patterns~\cite{han2026demaml}. Among them, Multi-GNN extends message passing to directed multigraphs~\cite{egressy2024directed}, and we evaluate its GIN variant, Multi-GIN; FraudGT incorporates transaction attributes through edge-aware attention~\cite{lin2024fraudgt}. This formulation is well aligned with benchmarks that provide transaction-level labels, but it assumes that supervision is available at the same granularity. \method considers the complementary setting in which labels are provided only at the account level.

\subsection{Multiple Instance Learning}
Multiple instance learning (MIL) learns from labeled bags whose constituent instances are not individually annotated~\cite{dietterich1997solving,maron1998framework}. Under the classical MIL assumption, a bag is positive if at least one of its instances is positive, motivating max-based aggregation of instance evidence~\cite{andrews2002support,carbonneau2018multiple}. Later work introduced smooth and attention-weighted alternatives that distribute learning signals across multiple instances~\cite{ilse2018attention,li2021dual}. MIL has since been applied to problems in which annotations are available at a coarser level than the evidence of interest, including whole-slide pathology~\cite{campanella2019clinical} and weakly supervised video anomaly detection~\cite{sultani2018real}. Account-level AML exhibits the same supervision structure: labels are assigned to accounts, while the evidence underlying those labels resides in their transactions. \method formulates each account-window as a bag and its transactions as instances, and constructs the account representation through dimension-wise max pooling over transaction embeddings.

\section{Methodology}
\label{sec:method}

\subsection{Problem Formulation}
Let $\mathcal{G}=(\mathcal{V},\mathcal{E})$ be a directed multigraph, where each node $u\in\mathcal{V}$ represents an account and each edge $e=(p,q)\in\mathcal{E}$ represents a transaction from account $p$ to account $q$. Each transaction carries an attribute vector $\mathbf{r}_e$, which includes its amount, timestamp, and other available information. Because multiple transactions may occur between the same pair of accounts, $\mathcal{G}$ is represented as a multigraph.

Let $\mathcal{W}$ partition the observation period into consecutive windows. For an account $u$ and a window $w\in\mathcal{W}$, we first define the complete set of incident transactions observed in that window as
\begin{equation}
\mathcal{E}^{\mathrm{raw}}_{u,w}
=\{e=(p,q)\in\mathcal{E}\mid u\in\{p,q\},\;t_e\in w\},
\label{eq:incident}
\end{equation}
where $t_e$ is the transaction timestamp. To bound memory, preprocessing first draws, for each account $u$, one fixed random subset of at most $K$ transactions from all transactions incident to $u$ over the full observation period. It then intersects this retained account-level subset with each temporal window $w$. We write $\mathcal{E}_{u,w}\subseteq\mathcal{E}^{\mathrm{raw}}_{u,w}$ for the resulting window-specific transaction set. The fixed account-level subset is shared across every method and training seed, and each window-specific set defines both the bag and its account-window label. Each non-empty pair $(u,w)$ forms one account-window sample, and the objective is to predict its label $y_{u,w}\in\{0,1\}$. The benchmark datasets also provide a transaction label $y_e\in\{0,1\}$ for each edge. These transaction labels are withheld during account-supervised training and are used only in the post-hoc ranking and controlled diagnostics.

To learn from account-window labels alone, \method maps each transaction incident to $u$ to an account-specific representation $\mathbf{x}_{u,e}$ and forms the bag
\begin{equation}
\mathcal{B}_{u,w}
=\{\mathbf{x}_{u,e}\mid e\in\mathcal{E}_{u,w}\}.
\label{eq:bag}
\end{equation}
This preserves individual transactions as instances while using the account-window as the unit of prediction.

The resulting bag follows the standard multiple instance relation
\begin{equation}
y_{u,w} \;=\; \max_{e\,\in\,\mathcal{E}_{u,w}} y_e ,
\label{eq:mi}
\end{equation}
under which an account-window is positive if and only if it contains at least one positive transaction. The model is trained using only $y_{u,w}$; the constituent labels $y_e$ never enter its objective.
Because each transaction is incident to two endpoint accounts, a positive transaction contributes a positive instance to both endpoint-specific bags. The resulting account-window label denotes incidence to a flagged transaction; it is a controlled supervision proxy and does not adjudicate that both endpoint accounts are culpable.

\begin{table*}[t]
\centering
\caption{Dataset statistics after each loader's tail cut. A sample is an account--window pair. Flagged accounts and transactions are counted over the full graph, whereas flagged test samples are counted in the test window. Parentheses give percentages of the corresponding totals. The last two columns indicate whether a dataset contains real transactions and is publicly available.}
\label{tab:datasets}
\small
\begin{tabular*}{\textwidth}{@{\extracolsep{\fill}}lrrrrrrcc@{}}
\toprule
Dataset & Accounts & Transactions
 & \makecell{Account-window\\samples}
 & \makecell{Flagged\\accounts}
 & \makecell{Flagged\\transactions}
 & \makecell{Flagged test\\samples}
 & \makecell{Real\\data}
 & \makecell{Publicly\\available} \\
\midrule
AMLworld~\cite{amlpatterns2024realistic} & 515,078 & 5,077,097 & 1,105,081 & 5,862 (1.14\%) & 1,522 (0.03\%) & 1,643 (0.46\%) & \xmark & \cmark \\
AMLNet~\cite{huda2025amlnet} & 11,000 & 1,088,410 & 33,000 & 1,468 (13.35\%) & 1,709 (0.16\%) & 247 (2.25\%) & \xmark & \cmark \\
AMLSim~\cite{suzumura2021amlsim} & 9,999 & 1,323,234 & 29,996 & 1,639 (16.39\%) & 1,719 (0.13\%) & 423 (4.23\%) & \xmark & \cmark \\
Elliptic++~\cite{elmougy2023demystifying} & 265,354 & 1,661,916 & 268,650 & 14,266 (5.38\%) & 15,863 (0.95\%) & 1,747 (5.24\%) & \cmark & \cmark \\
\hofinet & 452,816 & 4,732,130 & 758,994 & 9,644 (2.13\%) & 14,190 (0.31\%) & 2,723 (1.31\%) & \cmark & \xmark \\
\bottomrule
\end{tabular*}
\end{table*}

\subsection{Instance Construction}\label{subsec:instances}
Given an account $u$ and one of its incident transactions $e$, \method constructs the instance representation $\mathbf{x}_{u,e}$ from three sources of information. The first describes the transaction itself, including its log-transformed amount, direction, time of day, and channel. The second describes the behavior of account $u$ up to the time of the transaction, expressing the amount relative to its previous transfers, the time interval relative to its transaction rhythm, and the counterparty relative to its transaction history. The third describes the account at the other end of the transaction using the same historical features computed at the same point in time. Together, these features place each transaction in the behavioral context of both participating accounts. Rather than propagating information across neighboring nodes, \method attaches the counterparty context directly to the transaction connecting the two accounts, allowing each instance to be encoded independently. Table~\ref{tab:featurelist} lists them.

All historical features are computed using only transactions observed at or before the timestamp of the corresponding instance, thereby preventing information from future transactions from entering its representation. The consecutive windows in Equation~\ref{eq:incident} prevent a single bag from combining behavior across temporal split boundaries. Every resulting account-window bag has width at most $K$ and is padded when necessary.

\subsection{Account-Level Prediction}\label{subsec:pooling}
For each transaction $e$ associated with account $u$, a shared encoder $f_\theta$ maps the instance representation to a $d$-dimensional hidden embedding,
\begin{equation}
\mathbf{h}_{u,e}=f_\theta(\mathbf{x}_{u,e})\in\mathbb{R}^{d}.
\label{eq:encode}
\end{equation}
The same encoder is applied independently to every transaction in the bag. We implement $f_\theta$ as a two-layer perceptron with ReLU activations and dropout.

\method constructs the account representation by taking the maximum activation in each hidden dimension,
\begin{equation}
\mathbf{z}_{u,w}[j]
=\max_{e\in\mathcal{E}_{u,w}}\mathbf{h}_{u,e}[j],
\qquad j=1,\ldots,d.
\label{eq:pool}
\end{equation}
This operation treats the transactions as an unordered set and is invariant to their permutation. Because the maximum is taken separately in each hidden dimension, different dimensions of $\mathbf{z}_{u,w}$ may originate from different transactions; the pooled representation need not match any one transaction embedding.

A linear prediction head $g_\phi$ maps the pooled representation to the account-window logit $\ell_{u,w}$, and $\sigma$ denotes the sigmoid function that converts the logit into a prediction probability,
\begin{equation}
\ell_{u,w}=g_\phi(\mathbf{z}_{u,w}),
\qquad
\hat{y}_{u,w}=\sigma(\ell_{u,w}).
\label{eq:prediction}
\end{equation}
Training minimizes a weighted binary cross-entropy over the set $\mathcal{S}_{\mathrm{tr}}$ of training account-window pairs,
\begin{equation}
\mathcal{L}=-\frac{1}{|\mathcal{S}_{\mathrm{tr}}|}
\sum_{(u,w)\in\mathcal{S}_{\mathrm{tr}}}
\Big[\alpha\,y_{u,w}\log\hat{y}_{u,w}
+(1-y_{u,w})\log(1-\hat{y}_{u,w})\Big],
\label{eq:loss}
\end{equation}
where $\alpha$ is the ratio of negative to positive bags in the training set. Only account-level labels enter this objective. Under dimension-wise max pooling, the gradient in each hidden dimension is routed through the transaction that produced its maximum activation, allowing the model to determine which transactions support the account prediction without transaction-level supervision.

\subsection{Transaction Ranking and Attribution}
\method provides two complementary transaction-level quantities. The first assigns a candidate-ranking score to every transaction by applying the shared prediction head to its embedding,
\begin{equation}
s_{u,w,e}=g_\phi(\mathbf{h}_{u,e}).
\label{eq:instance_score}
\end{equation}
Although $s_{u,w,e}$ receives no transaction-level supervision, it provides a complete ranking of the transactions in $\mathcal B_{u,w}$. It measures how strongly a transaction would be classified on its own and is the score evaluated by the within-account ranking metrics. It is not a faithful decomposition of the account decision, because \method applies $g_\phi$ after pooling.

The second quantity provides that exact forward-path attribution. Let $\pi_{u,w}(j)=\arg\max_{e\in\mathcal{E}_{u,w}}\mathbf{h}_{u,e}[j]$ identify the transaction that supplies hidden dimension $j$, and let $\boldsymbol{\beta}$ and $b$ denote the weights and bias of $g_\phi$. The signed contribution of transaction $e$ is
\begin{equation}
c_{u,w,e}=\!\!\sum_{j\,:\,\pi_{u,w}(j)=e}\!\! \beta_j\,\mathbf{z}_{u,w}[j],
\qquad
\sum_{e\in\mathcal{E}_{u,w}}c_{u,w,e}=\ell_{u,w}-b.
\label{eq:credit}
\end{equation}
Unlike $s_{u,w,e}$, $c_{u,w,e}$ exactly decomposes the account-window logit excluding its bias, but it is sparse: a hidden representation of width $d$ assigns a non-zero contribution to at most $d$ transactions. We use $s_{u,w,e}$ for the complete review ranking and retain $c_{u,w,e}$ as a distinct attribution of the model's actual forward path; the two quantities answer different questions and are not treated as interchangeable explanations. This attribution describes the realized computation rather than an investigator-validated rationale.

\section{Experimental Setup}
\label{sec:experiments}

\subsection{Datasets}\label{subsec:setup}
We evaluate on four public benchmarks and one non-public real-world interbank network, summarized in Table~\ref{tab:datasets}.

\begin{itemize}[leftmargin=1.4em, itemsep=2pt, topsep=3pt, parsep=0pt]
\item \textbf{AMLworld}~\cite{amlpatterns2024realistic} is a synthetic banking-transaction benchmark developed to compare AML detection models using complete transaction-level laundering labels. We use its HI-Small release, which contains multiple simulated laundering typologies embedded in otherwise ordinary transactions.

\item \textbf{AMLNet}~\cite{huda2025amlnet} is generated by a knowledge-based multi-agent framework designed to model and detect laundering transactions. Its laundering covers the placement, layering and integration stages and typologies such as structuring.

\item \textbf{AMLSim}~\cite{suzumura2021amlsim} is a multi-agent simulator that generates banking transactions with known laundering patterns for testing machine-learning and graph-based detection algorithms. We use a fixed public snapshot rather than regenerating the data, so every method sees the same graph.

\item \textbf{Elliptic++}~\cite{elmougy2023demystifying} is a real Bitcoin dataset introduced for financial forensics involving both fraudulent transactions and illicit wallet nodes. We treat wallets as accounts and their incident Bitcoin transfers as transaction instances.

\item \textbf{\hofinet} is a non-public real-world interbank settlement network used for operational account monitoring. It tests whether results obtained on synthetic and cryptocurrency data transfer to an institutional setting.
\end{itemize}

Although the datasets were introduced for different detection problems, we convert all five into the same account-window classification task. An account-window sample is flagged when it contains at least one flagged transaction. The datasets do not provide independently adjudicated account-window case labels, so this coarsening emulates account supervision rather than reproducing the noise and reporting delay of operational case labels. \method and the account-level baselines are trained only with the resulting account-window labels, while the original transaction labels are withheld for transaction-ranking and controlled diagnostics. Multi-GIN and FraudGT use the transaction labels under their native supervision setting. This construction isolates supervision granularity under an exact bag-label relation, but independently assigned institutional case labels remain outside its scope.

\subsection{Baselines}
We organize the baselines by input structure and prediction granularity. The first three groups share \method's account-window supervision; the last uses transaction labels and predicts edges.
\begin{itemize}[leftmargin=1.4em, itemsep=2pt, topsep=3pt, parsep=0pt]
\item \textbf{Tabular classification} includes LR, a multilayer perceptron and XGBoost~\cite{xgboost2016} on aggregated account features alone.

\item \textbf{General graph node classification} includes GCN~\cite{kipf2016semi}, GAT~\cite{velivckovic2017graph}, GraphSAGE~\cite{hamilton2017inductive} and H2GCN~\cite{zhu2020beyond} as standard message-passing and het\-er\-oph\-i\-ly-aware ac\-count clas\-si\-fiers.

\item \textbf{Fraud-oriented graph node classification} includes CARE-GNN~\cite{dou2020enhancing}, PC-GNN~\cite{liu2021pick}, BWGNN~\cite{tang2022rethinking}, GAGA~\cite{wang2023label}, GCD-GNN~\cite{liu2025gcd} and GPA~\cite{qiu2025gpa}.

\item \textbf{Transaction-level AML classification} includes Multi-GIN~\cite{egressy2024directed} and FraudGT~\cite{lin2024fraudgt}. For ac\-count-lev\-el eval\-u\-a\-tion, an ac\-count-win\-dow receives the maximum predicted probability among its incident transactions.
\end{itemize}

The model families retain their native input structures. Tabular and graph node classifiers receive aggregate account features, graph models additionally receive window-specific account connectivity, and \method receives transaction-instance features with causal account and counterparty context. Table~\ref{tab:main} is therefore an end-to-end comparison of account-detection pipelines rather than a feature-matched estimate of the effect of pooling alone. The controlled variants in Section~\ref{subsec:controlled-setup} provide that feature-matched comparison by holding the transaction inputs, encoder and training protocol fixed.

\subsection{Experimental Protocol}
Transactions are ordered by time and split $60/20/20$ into training, validation and test windows. Each account retains a fixed subset of transactions capped at the dataset-specific $99$th percentile, up to $1{,}024$; this subset is shared across methods and training seeds. Transaction-ranking metrics are computed over these retained instances. Appendix~\ref{app:hyper} provides bag-size and retention details.

Each temporal window forms a separate graph for graph baselines, preventing message passing across split boundaries while allowing transductive use of within-window topology. Model, epoch and decision-threshold selection use validation data only: neural models early-stop after $30$ epochs without AP improvement, and XGBoost after $50$ rounds. Feature scaling is fitted on training rows. We report means and standard deviations over five training seeds under the fixed temporal cut; the deviations do not measure variation across alternative time periods. Appendix~\ref{app:hyper} gives the search grids.

\subsection{Controlled Aggregation Diagnostics}\label{subsec:controlled-setup}
We compare Pre-aggregation, MIL mean, Scalar max and \method while holding their transaction features, account-window labels, temporal split, encoder, training budget and search space fixed. Each fitted model is reused without retraining or threshold reselection in two controlled test-time interventions. For each training seed, we randomly order the ordinary transactions in every test bag once and share that order across all four variants. Nested prefixes expose $1$, $2$, $4$, $8$, $16$, $32$, $64$ and $128$ ordinary transactions. A shorter bag uses all available transactions, and Full restores the original retained bag. \emph{Positive dilution} retains every flagged transaction, progressively exposes ordinary transactions only in positive bags and reports account AP over the full test set while negative bags remain fixed. \emph{Negative exposure} keeps positive bags fixed, progressively exposes ordinary transactions in negative bags and reports FPR at the validation-selected threshold. Transaction labels are consulted only after training to distinguish flagged from ordinary transactions when constructing these diagnostic masks. They do not enter training, model selection or threshold selection.

\subsection{Implementation}
Experiments used one NVIDIA GB10 GPU with 121~GiB of unified memory on Ubuntu 24.04.4 LTS. We used Python 3.12.13, PyTorch 2.11.0 with CUDA 12.8, PyTorch Geometric 2.8.0, scikit-learn 1.9.0 and XGBoost 3.3.0.

\begin{table*}[t]
\centering
\caption{Detection on the test window over five seeds. Each cell reports the mean with its standard deviation on the second line. AP is average precision, AUC is AUROC, F1 is macro $F_1$, and GM is G-mean. Best per column in bold and second best underlined.}
\label{tab:main}
\footnotesize
\setlength{\tabcolsep}{1.2pt}
\begin{tabular*}{\textwidth}{@{\extracolsep{\fill}}lcccccccccccccccccccc@{}}
\toprule
\multirow{2}{*}{Method} & \multicolumn{4}{c}{AMLworld} & \multicolumn{4}{c}{AMLNet} & \multicolumn{4}{c}{AMLSim} & \multicolumn{4}{c}{Elliptic++} & \multicolumn{4}{c}{\hofinet} \\
\cmidrule(lr){2-5}\cmidrule(lr){6-9}\cmidrule(lr){10-13}\cmidrule(lr){14-17}\cmidrule(lr){18-21}
  & AP & AUC & F1 & GM & AP & AUC & F1 & GM & AP & AUC & F1 & GM & AP & AUC & F1 & GM & AP & AUC & F1 & GM \\
\midrule
\multicolumn{21}{l}{\textit{Tabular classification}} \\
LR & \score{0.062}{0.000} & \score{0.755}{0.000} & \score{0.552}{0.000} & \score{0.527}{0.000} & \score{0.373}{0.000} & \score{0.950}{0.000} & \score{0.706}{0.000} & \score{0.795}{0.000} & \score{0.357}{0.000} & \score{0.719}{0.000} & \score{0.688}{0.000} & \score{0.547}{0.000} & \score{0.252}{0.000} & \score{0.777}{0.000} & \score{0.053}{0.000} & \score{0.054}{0.000} & \score{0.539}{0.000} & \score{0.993}{0.000} & \score{0.813}{0.000} & \score{\textbf{0.984}}{0.000} \\
MLP & \score{0.091}{0.038} & \score{0.765}{0.032} & \score{0.574}{0.018} & \score{0.415}{0.025} & \score{0.590}{0.077} & \score{0.963}{0.024} & \score{0.772}{0.019} & \score{0.824}{0.061} & \score{0.326}{0.025} & \score{0.665}{0.018} & \score{0.664}{0.011} & \score{0.471}{0.020} & \score{0.208}{0.114} & \score{0.796}{0.076} & \score{0.050}{0.000} & \score{0.015}{0.006} & \score{0.564}{0.025} & \score{0.993}{0.000} & \score{0.807}{0.007} & \score{0.966}{0.010} \\
XGBoost & \score{0.136}{0.000} & \score{0.810}{0.000} & \score{0.595}{0.000} & \score{0.352}{0.000} & \score{0.699}{0.000} & \score{0.976}{0.000} & \score{0.810}{0.000} & \score{0.879}{0.000} & \score{\underline{0.839}}{0.000} & \score{\underline{0.944}}{0.000} & \score{\underline{0.945}}{0.000} & \score{\underline{0.904}}{0.000} & \score{0.232}{0.000} & \score{0.664}{0.000} & \score{\textbf{0.621}}{0.000} & \score{0.396}{0.000} & \score{\underline{0.648}}{0.000} & \score{\underline{0.995}}{0.000} & \score{0.818}{0.000} & \score{0.975}{0.000} \\
\midrule
\multicolumn{21}{l}{\textit{General graph node classification}} \\
GCN & \score{0.120}{0.022} & \score{0.767}{0.036} & \score{0.590}{0.012} & \score{0.395}{0.011} & \score{0.651}{0.012} & \score{0.969}{0.005} & \score{0.783}{0.016} & \score{0.838}{0.016} & \score{0.420}{0.045} & \score{0.880}{0.014} & \score{0.688}{0.014} & \score{0.628}{0.087} & \score{0.195}{0.007} & \score{0.780}{0.022} & \score{0.510}{0.011} & \score{0.156}{0.039} & \score{0.442}{0.018} & \score{0.974}{0.003} & \score{0.746}{0.008} & \score{0.779}{0.017} \\
GAT & \score{0.194}{0.037} & \score{0.820}{0.017} & \score{0.613}{0.036} & \score{0.546}{0.025} & \score{0.511}{0.136} & \score{0.929}{0.028} & \score{0.741}{0.035} & \score{0.797}{0.025} & \score{0.354}{0.117} & \score{0.828}{0.043} & \score{0.627}{0.077} & \score{0.620}{0.121} & \score{0.229}{0.134} & \score{0.812}{0.086} & \score{0.487}{0.002} & \score{0.013}{0.028} & \score{0.476}{0.059} & \score{0.981}{0.008} & \score{0.768}{0.030} & \score{0.794}{0.070} \\
GraphSAGE & \score{0.202}{0.007} & \score{0.869}{0.007} & \score{0.629}{0.005} & \score{0.491}{0.017} & \score{0.632}{0.013} & \score{0.965}{0.012} & \score{0.782}{0.011} & \score{0.840}{0.016} & \score{0.710}{0.059} & \score{0.932}{0.012} & \score{0.820}{0.031} & \score{0.718}{0.047} & \score{0.175}{0.092} & \score{0.729}{0.137} & \score{0.557}{0.046} & \score{0.475}{0.188} & \score{0.510}{0.021} & \score{0.988}{0.002} & \score{0.786}{0.009} & \score{0.899}{0.035} \\
H2GCN & \score{0.183}{0.043} & \score{0.796}{0.007} & \score{0.636}{0.019} & \score{0.455}{0.021} & \score{0.678}{0.059} & \score{0.970}{0.010} & \score{0.800}{0.031} & \score{0.841}{0.030} & \score{0.662}{0.051} & \score{0.929}{0.014} & \score{0.782}{0.031} & \score{0.691}{0.075} & \score{0.173}{0.076} & \score{0.793}{0.184} & \score{0.549}{0.068} & \score{0.404}{0.379} & \score{0.382}{0.131} & \score{0.967}{0.017} & \score{0.696}{0.065} & \score{0.784}{0.134} \\
\midrule
\multicolumn{21}{l}{\textit{Fraud-oriented graph node classification}} \\
CARE-GNN & \score{0.073}{0.007} & \score{0.710}{0.018} & \score{0.569}{0.002} & \score{0.339}{0.010} & \score{0.492}{0.013} & \score{0.928}{0.007} & \score{0.733}{0.005} & \score{0.750}{0.026} & \score{0.154}{0.024} & \score{0.744}{0.015} & \score{0.571}{0.006} & \score{0.559}{0.053} & \score{0.174}{0.133} & \score{0.694}{0.090} & \score{0.542}{0.113} & \score{0.262}{0.221} & \score{0.036}{0.006} & \score{0.668}{0.053} & \score{0.508}{0.012} & \score{0.478}{0.079} \\
PC-GNN & \score{0.043}{0.024} & \score{0.643}{0.027} & \score{0.545}{0.021} & \score{0.253}{0.071} & \score{0.146}{0.044} & \score{0.800}{0.022} & \score{0.589}{0.017} & \score{0.456}{0.041} & \score{0.169}{0.022} & \score{0.756}{0.025} & \score{0.432}{0.065} & \score{0.645}{0.031} & \score{\underline{0.331}}{0.182} & \score{0.831}{0.118} & \score{\underline{0.603}}{0.107} & \score{0.371}{0.228} & \score{0.360}{0.009} & \score{0.952}{0.014} & \score{0.718}{0.002} & \score{0.760}{0.020} \\
BWGNN & \score{0.175}{0.031} & \score{0.820}{0.024} & \score{0.624}{0.024} & \score{0.529}{0.017} & \score{0.672}{0.022} & \score{0.959}{0.020} & \score{0.808}{0.006} & \score{0.836}{0.037} & \score{0.415}{0.244} & \score{0.867}{0.081} & \score{0.591}{0.177} & \score{0.599}{0.132} & \score{0.287}{0.176} & \score{\underline{0.858}}{0.060} & \score{0.592}{0.056} & \score{\underline{0.573}}{0.188} & \score{0.329}{0.060} & \score{0.971}{0.003} & \score{0.683}{0.013} & \score{0.834}{0.036} \\
GAGA & \score{0.036}{0.011} & \score{0.716}{0.020} & \score{0.543}{0.012} & \score{0.437}{0.070} & \score{0.478}{0.018} & \score{0.878}{0.028} & \score{0.745}{0.011} & \score{0.778}{0.021} & \score{0.329}{0.060} & \score{0.776}{0.014} & \score{0.654}{0.018} & \score{0.552}{0.056} & \score{0.187}{0.026} & \score{0.819}{0.021} & \score{0.430}{0.155} & \score{\textbf{0.664}}{0.157} & \score{0.325}{0.037} & \score{0.969}{0.008} & \score{0.700}{0.020} & \score{0.777}{0.064} \\
GCD-GNN & \score{0.212}{0.035} & \score{0.855}{0.024} & \score{0.638}{0.026} & \score{0.492}{0.016} & \score{0.705}{0.068} & \score{0.977}{0.009} & \score{0.816}{0.036} & \score{0.842}{0.032} & \score{0.548}{0.072} & \score{0.872}{0.027} & \score{0.711}{0.055} & \score{0.538}{0.089} & \score{0.077}{0.009} & \score{0.654}{0.027} & \score{0.251}{0.221} & \score{0.090}{0.159} & \score{0.461}{0.042} & \score{0.976}{0.015} & \score{0.767}{0.014} & \score{0.847}{0.047} \\
GPA & \score{0.187}{0.036} & \score{0.862}{0.004} & \score{0.621}{0.014} & \score{0.519}{0.029} & \score{0.761}{0.015} & \score{0.983}{0.001} & \score{0.828}{0.011} & \score{0.876}{0.019} & \score{0.751}{0.037} & \score{0.942}{0.006} & \score{0.854}{0.035} & \score{0.812}{0.067} & \score{0.112}{0.010} & \score{0.692}{0.045} & \score{0.533}{0.014} & \score{0.269}{0.059} & \score{0.615}{0.008} & \score{0.994}{0.000} & \score{0.814}{0.003} & \score{0.944}{0.009} \\
\midrule
\multicolumn{21}{l}{\textit{Transaction-level AML classification}} \\
Multi-GIN & \score{0.326}{0.073} & \score{\underline{0.944}}{0.006} & \score{0.670}{0.035} & \score{\textbf{0.668}}{0.019} & \score{\textbf{0.963}}{0.010} & \score{\textbf{0.999}}{0.000} & \score{\textbf{0.947}}{0.016} & \score{\textbf{0.961}}{0.015} & \score{0.717}{0.079} & \score{0.906}{0.030} & \score{0.856}{0.044} & \score{0.774}{0.056} & \score{0.241}{0.064} & \score{0.845}{0.039} & \score{0.565}{0.022} & \score{0.297}{0.033} & \score{0.558}{0.080} & \score{0.994}{0.002} & \score{\textbf{0.826}}{0.019} & \score{\underline{0.975}}{0.016} \\
FraudGT & \score{\underline{0.432}}{0.024} & \score{0.933}{0.009} & \score{\underline{0.734}}{0.013} & \score{\underline{0.596}}{0.037} & \score{0.915}{0.011} & \score{0.985}{0.002} & \score{0.925}{0.005} & \score{0.945}{0.012} & \score{0.107}{0.006} & \score{0.627}{0.010} & \score{0.505}{0.020} & \score{0.236}{0.215} & \score{0.069}{0.005} & \score{0.707}{0.024} & \score{0.493}{0.002} & \score{0.075}{0.027} & \score{0.543}{0.014} & \score{0.987}{0.001} & \score{0.795}{0.009} & \score{0.887}{0.036} \\
\midrule
\multicolumn{21}{l}{\textit{Ours}} \\
\method & \score{\textbf{0.467}}{0.013} & \score{\textbf{0.952}}{0.005} & \score{\textbf{0.739}}{0.007} & \score{0.593}{0.012} & \score{\underline{0.929}}{0.002} & \score{\underline{0.993}}{0.001} & \score{\underline{0.934}}{0.002} & \score{\underline{0.947}}{0.007} & \score{\textbf{0.998}}{0.001} & \score{\textbf{1.000}}{0.001} & \score{\textbf{0.994}}{0.002} & \score{\textbf{0.989}}{0.003} & \score{\textbf{0.433}}{0.036} & \score{\textbf{0.937}}{0.005} & \score{0.597}{0.022} & \score{0.356}{0.042} & \score{\textbf{0.672}}{0.011} & \score{\textbf{0.995}}{0.000} & \score{\underline{0.823}}{0.005} & \score{0.900}{0.010} \\
\bottomrule
\end{tabular*}
\end{table*}

\subsection{Evaluation Metrics}
We report average precision (AP), area under the receiver operating characteristic curve (AUROC), macro $F_1$ and G-mean. For a decision threshold selected on the validation set, let $\mathrm{TP}$, $\mathrm{TN}$, $\mathrm{FP}$ and $\mathrm{FN}$ denote the entries of the test-set confusion matrix. We define the true-positive, true-negative and false-positive rates as $\mathrm{TPR}=\mathrm{TP}/(\mathrm{TP}+\mathrm{FN})$, $\mathrm{TNR}=\mathrm{TN}/(\mathrm{TN}+\mathrm{FP})$ and $\mathrm{FPR}=\mathrm{FP}/(\mathrm{FP}+\mathrm{TN})$.

AP and AUROC evaluate the ranking induced by the predicted account scores without fixing a decision threshold~\cite{davis2006relationship,fawcett2006introduction}. Let $M$ be the number of distinct score thresholds, and let $P_m$ and $R_m$ be precision and recall at the $m$th threshold, ordered so that recall increases. We compute
\begin{equation}
\begin{aligned}
\mathrm{AP} &= \sum_{m=1}^{M}(R_m-R_{m-1})P_m,\\
\mathrm{AUROC} &= \int_{0}^{1}\mathrm{TPR}(r)\,d r,
\end{aligned}
\label{eq:ranking_metrics}
\end{equation}
where $r$ denotes FPR along the ROC curve. AP is our primary metric because AML monitoring produces a ranked list of alerts under limited investigative capacity: a useful model must recover suspicious accounts while limiting the legitimate accounts sent for review. AP summarizes this precision--recall trade-off across operating points and is likewise used in an operational AML evaluation to summarize performance across alert target rates~\cite{bis2025hertha}. Methods are compared by a two-sided $95\%$ Student-$t$ interval on the five
per-seed differences in AP against the strongest account-supervised baseline,
which covers training variation under the fixed temporal split, not
sampling variation over accounts.

Macro $F_1$ and G-mean evaluate the binary predictions obtained at the validation-selected threshold. Following their standard definitions~\cite{sokolova2009systematic,kubat1997addressing}, we compute
\begin{equation}
\begin{aligned}
\mathrm{Macro}\text{-}F_1
&=\frac{1}{2}\left(
\frac{2\mathrm{TP}}{2\mathrm{TP}+\mathrm{FP}+\mathrm{FN}}
+\frac{2\mathrm{TN}}{2\mathrm{TN}+\mathrm{FP}+\mathrm{FN}}
\right),\\
\mathrm{G\text{-}mean} &= \sqrt{\mathrm{TPR}\cdot\mathrm{TNR}}.
\end{aligned}
\label{eq:threshold_metrics}
\end{equation}
Macro $F_1$ assigns equal importance to the two classes, while G-mean measures whether sensitivity is maintained on both classes. We include them because $F_1$- and G-mean-based evaluation is common in graph-based AML and fraud detection~\cite{egressy2024directed,liu2021pick,tang2022rethinking}.

For within-account transaction ranking, let $\mathcal P$ be the positive test bags, let $\rho_i(r)$ denote the transaction ranked at position $r$ in bag $i$ by $s_{u,w,e}$, and let $m_i$ be the number of flagged transactions in that bag. We compute
\begin{equation}
\begin{aligned}
\mathrm{Hit@}k
&=\frac{1}{|\mathcal P|}\sum_{i\in\mathcal P}
\mathbb I\!\left[\sum_{r=1}^{\min(k,n_i)}y_{\rho_i(r)}>0\right],\\
\mathrm{AP}_i
&=\frac{1}{m_i}\sum_{r=1}^{n_i}
\mathrm{Prec}_i(r)y_{\rho_i(r)},
\qquad
\mathrm{MAP}=\frac{1}{|\mathcal P_{\mathrm{mix}}|}
\sum_{i\in\mathcal P_{\mathrm{mix}}}\mathrm{AP}_i,
\end{aligned}
\label{eq:ranking_localization}
\end{equation}
where $n_i$ is the number of transactions in bag $i$, $\mathrm{Prec}_i(r)$ is precision through rank $r$, and $\mathcal P_{\mathrm{mix}}\subseteq\mathcal P$ contains the positive bags with both flagged and ordinary transactions. Hit@$k$ measures whether reviewing the first $k$ candidates exposes any flagged transaction. MAP measures the complete within-bag ranking and excludes all-flagged bags, for which every ordering is trivially perfect.

\section{Results}
\label{sec:results}

\subsection{Overall Detection Performance}\label{subsec:overall-results}

\begin{figure*}[t]
\centering
\includegraphics[width=\textwidth]{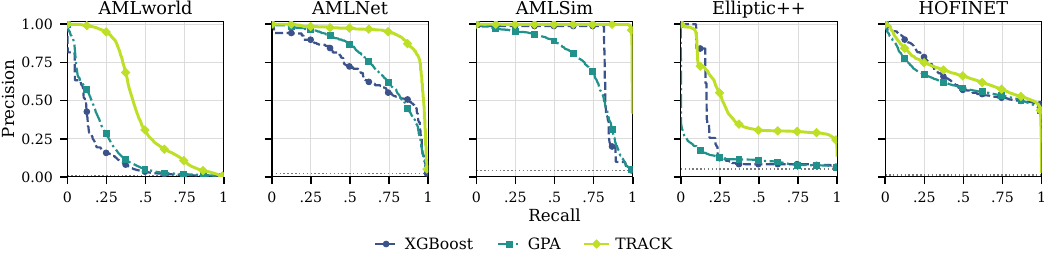}
\caption{Precision against recall on the test window, averaged over five seeds.
The opponents are XGBoost and GPA, the two account-supervised baselines with the highest average
precision over the five datasets, so every panel compares the same three
rankers. Dotted lines mark the prevalence, which is what a random ranker
attains. Table~\ref{tab:main} gives the areas.}
\label{fig:pr}
\end{figure*}

Table~\ref{tab:main} compares the complete account-detection pipelines under their respective input structures. Across all five datasets, \method leads every baseline trained with account-level supervision. The best competing family varies between tabular and graph models, so the result is not explained by the weakness of one particular architecture. Because these families differ in both representation and architecture, this table establishes end-to-end performance rather than attributing the difference to pooling alone. Importantly, the account-supervised result also holds on \hofinet, the non-public real-world interbank network, so the observed performance is not confined to public synthetic or cryptocurrency benchmarks. It excludes zero on four of the five datasets; on Elliptic++ the baseline's own seed spread of $\pm.182$ exceeds the $0.102$ gap.

AP compresses the precision--recall trade-off into a single value. Figure~\ref{fig:pr} therefore compares \method with representative account-supervised aggregate and graph baselines across the full recall range. Its advantage over these references generally persists across operating points, indicating that it is not produced by a narrow threshold choice and is relevant when an institution changes how many accounts it can review. \method does not dominate every threshold-dependent metric, indicating that its consistent advantage lies primarily in ranking suspicious accounts rather than in producing a universally optimal decision threshold. Because AML teams operate under a fixed review capacity, Table~\ref{tab:budget} reports Recall@$100$, Recall@$500$ and Recall@$1{,}000$, defined as the fraction of suspicious accounts recovered within each alert budget. At $100$ alerts the methods are close, with no gap above $0.020$. At $500$ and
$1{,}000$ \method recovers more suspicious accounts than the strongest
account-level baseline on four of the five datasets, with \hofinet as the only
exception.

\begin{table}[t]
\centering
\caption{Recall at a fixed alert budget, averaged over five seeds. The baseline
column holds the highest value among the account-level baselines at that budget.}
\label{tab:budget}
\small
\setlength{\tabcolsep}{5pt}
\begin{tabular}{@{}lcccccc@{}}
\toprule
\multirow{2}{*}{Dataset} & \multicolumn{2}{c}{Recall@$100$} & \multicolumn{2}{c}{Recall@$500$} & \multicolumn{2}{c}{Recall@$1{,}000$} \\
\cmidrule(lr){2-3}\cmidrule(lr){4-5}\cmidrule(lr){6-7}
& \method & Best & \method & Best & \method & Best \\
\midrule
AMLworld   & \textbf{0.061} & 0.056 & \textbf{0.280} & 0.171 & \textbf{0.383} & 0.229 \\
AMLNet     & \textbf{0.394} & 0.374 & \textbf{0.972} & 0.923 & \textbf{0.984} & 0.960 \\
AMLSim     & \textbf{0.236} & \textbf{0.236} & \textbf{0.998} & 0.823 & \textbf{0.999} & 0.851 \\
Elliptic++ & 0.056 & \textbf{0.057} & \textbf{0.193} & 0.165 & \textbf{0.271} & 0.231 \\
\hofinet   & \textbf{0.035} & \textbf{0.035} & 0.150 & \textbf{0.161} & 0.271 & \textbf{0.276} \\
\bottomrule
\end{tabular}
\end{table}

The transaction-supervised methods provide a complementary test of task alignment. They produce the highest account AP on some datasets but not on others, despite access to finer-grained labels, because their learning objective targets individual edges and the account prediction is formed only afterward. The results therefore separate supervision granularity from prediction granularity: transaction labels can improve transaction recognition without guaranteeing the strongest account ranking.

\begin{figure*}[t]
\centering
\includegraphics[width=\textwidth]{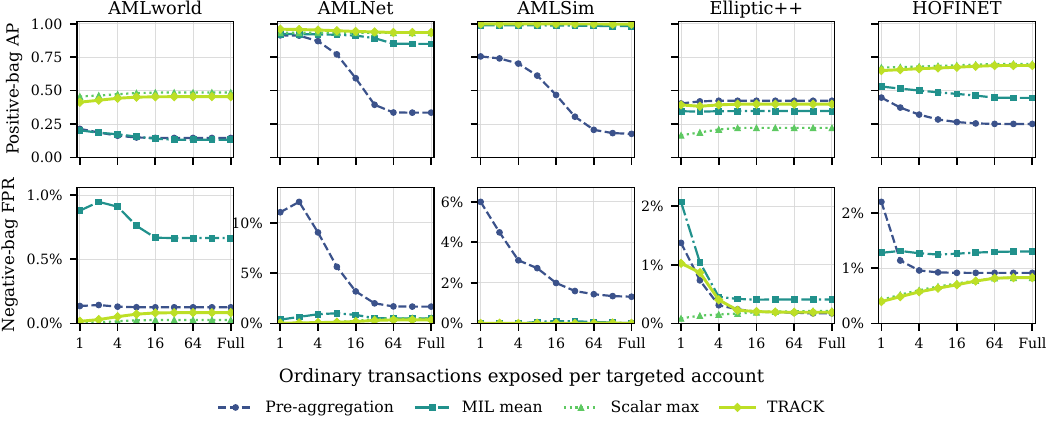}
\caption{Controlled evaluation of aggregation behavior, averaged over five seeds. Top: account AP as ordinary transactions are progressively added to positive bags while negative bags remain fixed. Bottom: false-positive rate at the validation-selected threshold as ordinary transactions are progressively added to negative bags while positive bags remain fixed. All models are trained once on the original bags and reused without retraining. Full denotes the original retained bag. Bottom-row axis ranges vary by dataset for readability.}
\label{fig:controlled-aggregation}
\end{figure*}

\subsection{Within-Account Transaction Ranking}\label{subsec:transaction-ranking}

\begin{table}[t]
\centering
\caption{Within-account transaction ranking on positive test bags. \method uses account labels, Multi-GIN and FraudGT use transaction labels, and Random ranks transactions uniformly. Results are averaged over five seeds. Best results are bold and second-best results are underlined.}
\label{tab:instance}
\footnotesize
\begin{tabular*}{\columnwidth}{@{\extracolsep{\fill}}llccc@{}}
\toprule
Dataset & Method & Hit@1 & Hit@5 & MAP \\
\midrule
\multirow{4}{*}{AMLworld}
  & Random    & 0.329 & 0.765 & 0.423 \\
  & \method    & \underline{0.762} & \underline{0.943} & \underline{0.809} \\
  & Multi-GIN  & \textbf{0.791} & \textbf{0.964} & \textbf{0.818} \\
  & FraudGT    & 0.341 & 0.761 & 0.425 \\
\midrule
\multirow{4}{*}{AMLNet}
  & Random    & 0.053 & 0.235 & 0.133 \\
  & \method    & \underline{0.976} & \underline{0.977} & \underline{0.931} \\
  & Multi-GIN  & \textbf{0.998} & \textbf{1.000} & \textbf{0.997} \\
  & FraudGT    & 0.059 & 0.246 & 0.139 \\
\midrule
\multirow{4}{*}{AMLSim}
  & Random    & 0.038 & 0.192 & 0.111 \\
  & \method    & \textbf{0.995} & \textbf{0.999} & \textbf{0.981} \\
  & Multi-GIN  & \underline{0.743} & \underline{0.789} & \underline{0.682} \\
  & FraudGT    & 0.039 & 0.182 & 0.113 \\
\midrule
\multirow{4}{*}{Elliptic++}
  & Random    & 0.901 & 0.993 & 0.656 \\
  & \method    & \underline{0.927} & \textbf{0.999} & \underline{0.785} \\
  & Multi-GIN  & \textbf{0.952} & \underline{0.998} & \textbf{0.859} \\
  & FraudGT    & 0.903 & 0.987 & 0.708 \\
\midrule
\multirow{4}{*}{\hofinet}
  & Random    & 0.463 & 0.731 & 0.373 \\
  & \method    & \textbf{0.843} & \textbf{0.983} & \textbf{0.842} \\
  & Multi-GIN  & \underline{0.819} & \underline{0.975} & \underline{0.822} \\
  & FraudGT    & 0.468 & 0.732 & 0.378 \\
\bottomrule
\end{tabular*}
\end{table}

Table~\ref{tab:instance} asks whether a model trained only to identify suspicious accounts can also rank useful transaction candidates for review. We order each positive account-window by $s_{u,w,e}$ and evaluate that ordering with transaction labels withheld from \method's training. As clarified in Section~\ref{sec:method}, this instance score is a complete candidate ranking rather than an exact decomposition of the account logit. Fraud-oriented node classifiers are omitted because they provide no native transaction score. Random, Multi-GIN and FraudGT provide the available ranking references.

\method ranks flagged transactions well above chance and remains competitive with methods trained directly from transaction labels, although neither supervision setting dominates across the public datasets. Operationally, its account alert can therefore be accompanied by a short list of candidate transactions even when transaction annotations are unavailable for training. This result establishes ranking utility rather than faithfulness to the pooled account logit. The separate contribution $c_{u,w,e}$ in Equation~\ref{eq:credit} provides the exact forward-path attribution. Elliptic++ is less diagnostic here because many positive bags hold no ordinary transaction, so almost any ordering succeeds.

FraudGT sharpens the same separation from the other side. On AMLworld it holds the
second-highest account average precision yet ranks at chance inside the bag, and its
margin over a uniform draw at hit@$1$ does not exceed $0.012$ on any dataset it covers.
Transaction supervision therefore does not by itself deliver a usable ordering within
an account.

\subsection{Preserving Sparse Positive Evidence}\label{subsec:positive-dilution}
\begin{table}[t]
\centering
\caption{Aggregation variants used in the controlled analysis, each trained with the same account-window labels and running from transaction inputs to an account logit.}
\label{tab:aggregation-variants}
\footnotesize
\setlength{\tabcolsep}{6pt}
\renewcommand{\arraystretch}{1.02}
\begin{tabular}{@{}L{0.23\columnwidth}L{0.68\columnwidth}@{}}
\toprule
Variant & Computation and interpretation \\
\midrule
Pre-aggregation
& $\mathrm{Mean}\rightarrow f_\theta\rightarrow g_\phi$\newline
Raw transaction features are averaged before encoding. \\
\addlinespace[2pt]
MIL mean
& $f_\theta\rightarrow\mathrm{Mean}\rightarrow g_\phi$\newline
Encoded transactions contribute uniformly. \\
\addlinespace[2pt]
Scalar max
& $f_\theta\rightarrow g_\phi\rightarrow\mathrm{Max}$\newline
One complete transaction score determines the account logit. \\
\addlinespace[2pt]
\method
& $f_\theta\rightarrow\mathrm{Feature\text{-}wise\ Max}\rightarrow g_\phi$\newline
Different transactions may supply different hidden dimensions. \\
\bottomrule
\end{tabular}
\end{table}

\begin{table}[t]
\centering
\caption{Account average precision at the full label budget for four pooling
operators sharing \method's encoder, features and grid, averaged over five seeds.
Best per row in bold and second best underlined.}
\label{tab:pooling}
\small
\begin{tabular*}{\columnwidth}{@{\extracolsep{\fill}}lcccc@{}}
\toprule
Dataset & MIL mean & Attention & Scalar max & \method \\
\midrule
AMLworld   & 0.137 & \underline{0.467} & \textbf{0.482} & \underline{0.467} \\
AMLNet     & 0.820 & \textbf{0.930} & 0.529 & \underline{0.929} \\
AMLSim     & 0.989 & 0.996 & \textbf{0.999} & \underline{0.998} \\
Elliptic++ & 0.358 & \underline{0.364} & 0.234 & \textbf{0.433} \\
\hofinet   & 0.443 & 0.661 & \textbf{0.693} & \underline{0.672} \\
\bottomrule
\end{tabular*}
\end{table}

Table~\ref{tab:aggregation-variants} summarizes the four controlled variants and exposes the two choices under study: whether aggregation occurs before or after representation learning, and whether it combines transactions uniformly or selectively. The top row of Figure~\ref{fig:controlled-aggregation} then reports the positive-dilution diagnostic defined in Section~\ref{subsec:controlled-setup}. Pre-aggregation deteriorates sharply on AMLNet and AMLSim, and both mean-based variants are substantially weaker on AMLworld. By contrast, scalar max and \method remain stable as ordinary transactions accumulate on the three datasets with larger or more variable bags. A summary formed too early can therefore erase a rare transaction pattern before supervision can shape its representation, while selective aggregation after instance encoding preserves that evidence.

The comparison between the two selective variants is more nuanced. Neither scalar max nor feature-wise max dominates across all datasets, so the diagnostic supports selective post-encoding aggregation as a family rather than attributing every gain uniquely to one max operator. Elliptic++ provides a useful boundary case because its short, concentrated bags leave little ordinary activity with which to dilute the positive evidence; the small changes there are therefore consistent with the mechanism tested by the intervention.

Table~\ref{tab:pooling} adds a learned operator under the same encoder and grid. Attention pooling~\cite{ilse2018attention} performs similarly to \method on most datasets but trails it on Elliptic++. Scalar max leads on AMLworld, AMLSim and \hofinet but drops sharply on AMLNet and Elliptic++. Across datasets, \method remains competitive without exhibiting the pronounced drops observed for Scalar max.

\subsection{False Alerts under Ordinary Activity}\label{subsec:negative-exposure}
The bottom row reports the negative-exposure diagnostic defined in Section~\ref{subsec:controlled-setup}. It examines the complementary risk that a selective operator could preserve suspicious evidence but also produce a false alert when a negative account presents more opportunities for a spuriously high response.

Because both the fitted parameters and the validation-selected threshold remain fixed along each curve, any change reflects sensitivity to added ordinary activity rather than retraining or threshold adaptation.

The selective variants maintain a low false-positive rate across the tested exposure levels and do not exhibit the monotonic increase that a simple multiple-opportunity account would predict. The larger changes occur in the mean-based variants: their scores can shift substantially when a small partial bag is replaced by a fuller account history, revealing sensitivity to bag composition rather than protection from false alerts alone. Read together, the two rows expose the relevant trade-off. Mean aggregation can suppress isolated responses, but it may also dilute sparse positive evidence and change calibration with bag size. Selective post-encoding aggregation preserves the evidence without producing a corresponding growth in false alerts. This controlled sensitivity check complements rather than repeats the detection comparison.

\subsection{Discussion}\label{subsec:discussion}
Account detection and within-account transaction ranking measure different properties of a model. Account AP evaluates how well complete account-windows are ordered after their transaction information has been aggregated, whereas Hit@$k$ and MAP evaluate the ordering of transactions within account-windows that are already known to be positive. A transaction-supervised method can therefore identify suspicious transfers accurately without producing the best account ranking after its edge scores are aggregated. Conversely, \method can produce a strong account ranking even when its candidate transaction score is not the strongest within every positive bag, because its account decision combines hidden features before classification. The two evaluations should consequently be read as complementary: Table~\ref{tab:main} measures case prioritization, while Table~\ref{tab:instance} measures how effectively each prioritized case can be narrowed to a short transaction list.

The controlled analysis also separates the broader benefit of selective
aggregation from the specific choice of feature-wise max pooling. Scalar max
and \method both avoid averaging every encoded transaction, which is why both
stay stable under positive dilution, but their decisions differ in structure:
scalar max needs one transaction to carry the whole account score, while
\method lets different transactions supply different hidden dimensions to
one decision. Feature-wise pooling therefore helps when an account
decision rests on complementary cues spread across transactions, which
Equation~\ref{eq:credit} traces back to the transfers that supplied them.

Finally, the account-window labels used here are controlled benchmark proxies rather than independently adjudicated institutional cases. Coarsening transaction labels makes the MIL relation in Equation~\ref{eq:mi} exact and allows supervision granularity to be compared without ambiguity, but it omits several properties of operational labels. A real case may be opened because of a combination of transfers, contextual information outside the transaction graph, delayed reporting, or investigator judgment, and may therefore violate the exact max relation. Our results establish that transaction-preserving learning is effective when fine-grained flags are hidden behind this coarser benchmark label; they do not establish that every institutional case can be reduced to one flagged transfer. Independently assigned case labels and analyst-reviewed rationales are therefore needed before the observed ranking and attribution count as operational evidence.

\section{Conclusion}
This study gives multiple instance learning a structural role in account-level
AML. Aggregate-first models collapse transactions before supervision can
determine which of them matter, and the controlled interventions show the cost:
mean-based aggregation loses positive evidence as ordinary activity
accumulates, while selective post-encoding aggregation holds that evidence
without a systematic increase in false-positive rates under the controlled
intervention. Across four public datasets and the non-public
\hofinet interbank network, \method obtains the highest account AP among the baselines
trained with account labels, and its alerts carry a transaction ranking
learned without transaction supervision. Both results rest on a label built by
coarsening transaction flags, so they establish the formulation under controlled weak supervision, not in deployment. Labels carrying reporting delay and
investigator judgment, where the bag-label identity is only approximate, are
the next test of the formulation.

\appendix

\section*{Appendix}

\section{Hyper-parameters}\label{app:hyper}
Table~\ref{tab:hyper} lists \method's settings. The graph baselines use two
layers at width $64$ and search learning rates $10^{-2}$ and
$10^{-3}$~\cite{dou2020enhancing,liu2021pick,tang2022rethinking}; depth,
patience and weight decay are common across the neural methods. XGBoost uses
$500$ trees of depth $6$ at shrinkage $0.05$ and logistic regression balanced
class weights, so Table~\ref{tab:main} compares implemented pipelines under
method-specific search spaces. The controlled aggregation study gives all four
variants an identical space.

\paragraph{Bag-size control.}
Median bag sizes are $2$ transactions on \hofinet and Elliptic++, $6$ on AMLworld, $189$ on AMLNet and $216$ on AMLSim. The dataset-specific caps range from $101$ to $1{,}024$. Each retained subset is drawn once with a fixed preprocessing seed and is not resampled by epoch. Under uniform random sampling, a post-hoc hypergeometric estimate indicates that $96.9$--$100.0\%$ of positive accounts retain at least one flagged transaction across the five datasets. Changing the cap from the $99$th percentile to the $95$th or $99.9$th changed
mean AP by at most $0.006$ on AMLNet and AMLSim.

\begin{table}[h]
\centering
\caption{\method's search grid and fixed settings. Searched values are separated by a comma; a single value was not searched.}
\label{tab:hyper}
\small
\renewcommand{\arraystretch}{0.95}
\setlength{\tabcolsep}{5pt}
\begin{tabular*}{\columnwidth}{@{\extracolsep{\fill}}L{0.25\columnwidth}L{0.68\columnwidth}@{}}
\toprule
Setting & Value \\
\midrule
Learning rate      & $3{\cdot}10^{-4}$, $10^{-3}$ \\
Hidden width       & $64$, $128$ \\
Encoder            & 2-layer perceptron, ReLU, dropout $0.2$ \\
Optimizer          & Adam, weight decay $10^{-4}$ \\
Epochs             & $200$, early stopping at patience $30$ \\
Batch              & $128$ bags \\
Class weight       & negative-to-positive ratio of the training bags \\
Bag width cap      & $99$th percentile of bag size, at most $1{,}024$ \\
Selection          & validation average precision \\
\bottomrule
\end{tabular*}
\end{table}

\section*{Ethical Considerations}

\method is intended to support account-level triage rather than make autonomous
enforcement decisions. False-positive predictions may subject legitimate
customers to unnecessary investigation or service restrictions, while false
negatives may leave illicit activity undetected. Transaction rankings and
model attributions should therefore support analyst review rather than be
treated as proof of wrongdoing.

The public datasets used in this study are synthetic or previously released
research datasets, while the non-public dataset is handled under its applicable
data-sharing restrictions. No personally identifiable information is released.
Because synthetic transaction labels and account constructions may not reflect
the noise, reporting delay, or demographic disparities of operational AML
systems, deployment would require institution-specific validation, fairness
auditing, privacy safeguards, and human oversight.

Releasing detailed detection methods may also inform adversarial evasion.
Accordingly, the model should be deployed with access controls, monitoring,
periodic reassessment, and complementary rule- and investigator-based review.

\bibliographystyle{ACM-Reference-Format}
\bibliography{aml}
\end{document}